\documentclass[11pt,a4paper]{article}
\usepackage{times,latexsym}
\usepackage{url}
\usepackage[T1]{fontenc}

\usepackage[acceptedWithA]{tacl2018v2}
\usepackage{xspace,mfirstuc,tabulary}
\usepackage{amsmath,bm}
\usepackage{multirow}
\usepackage{color}
\usepackage{multicol}
\usepackage{tabularx}
\usepackage{booktabs}
\usepackage{pbox}
\usepackage{framed}
\usepackage{array}
\usepackage{enumitem}
\usepackage{chngpage}
\usepackage{threeparttable}
\usepackage{dsfont}
\usepackage{graphicx}
\usepackage{enumitem}
\usepackage{epigraph}
\usepackage{CJKutf8}
\usepackage{amsfonts}
\usepackage{arydshln}

\usepackage{subfig}

\newif\iftaclinstructions
\taclinstructionsfalse %
\iftaclinstructions
\renewcommand{\confidential}{}
\renewcommand{\anonsubtext}{(No author info supplied here, for consistency with
TACL-submission anonymization requirements)}
\newcommand{\instr}
\fi

\iftaclpubformat %

\else

\fi

\newcommand{\eg}{{e.g.}}
\newcommand{\ie}{{i.e.}}
\newcommand{\vs}{{vs. }}
\newcommand{\dn}{{C$^3$}} %
\newcommand{\dnm}{${\text C}_{\text M}^{3}$}
\newcommand{\dnd}{${\text C}_{\text D}^{3}$}
\newcommand{\hhide}[1]{}

\title{Investigating Prior Knowledge for \\Challenging Chinese Machine Reading Comprehension}

\author{
  Kai Sun\textsuperscript{1}\Thanks{Part of this work was conducted when K. S. was an intern at the Tencent AI Lab, Bellevue, WA.}~~ Dian Yu\textsuperscript{2} ~~ Dong Yu\textsuperscript{2} ~~ Claire Cardie\textsuperscript{1}\\
 \textsuperscript{1}Cornell University, Ithaca, NY \\
 \textsuperscript{2}Tencent AI Lab, Bellevue, WA\\
   ks985@cornell.edu, \{yudian, dyu\}@tencent.com, cardie@cs.cornell.edu \\
}

\date{}

\begin{document}
\maketitle
\begin{CJK*}{UTF8}{gkai}

\begin{abstract}

Machine reading comprehension tasks require a machine reader to answer questions relevant to the given document. In this paper, we present the first free-form multiple-Choice Chinese machine reading Comprehension dataset ({\dn}), containing 13,369 documents (dialogues or more formally written mixed-genre texts) and their associated 19,577 multiple-choice free-form questions collected from Chinese-as-a-second-language examinations.

We present a comprehensive analysis of the prior knowledge (\ie, linguistic, domain-specific, and general world knowledge) needed for these real-world problems. We implement rule-based and popular neural methods and find that there is still a significant performance gap between the best performing model ($68.5\%$) and human readers ($96.0\%$), especially on problems that require prior knowledge. We further study the effects of distractor plausibility and data augmentation based on translated relevant datasets for English on model performance. We expect {\dn} to present great challenges to existing systems as answering $86.8\%$ of questions requires both knowledge within and beyond the accompanying document, and we hope that {\dn} can serve as a platform to study how to leverage various kinds of prior knowledge to better understand a given written or orally oriented text. {\dn} is available at \url{https://dataset.org/c3/}.

\end{abstract}

\section{Introduction}

\epigraph{\textit{``Language is, at best, a means of directing others to construct similar-thoughts from their own prior knowledge.''}}{\newcite{adams1982background}}

Machine reading comprehension (MRC) tasks have attracted substantial attention from both academia and industry. These tasks require a machine reader to answer questions relevant to a given document provided as input~\cite{poon2010machine,richardson2013mctest}. In this paper, we focus on \emph{\textbf{free-form multiple-choice}} MRC tasks --- given a document, select the correct answer option from all options associated with a free-form question, which is not limited to a single question type such as cloze-style questions formed by removing a span or a sentence in a text~\cite{hill2015goldilocks,bajgar2016embracing,mostafazadeh2016corpus,xie2018large,zheng2019chid} or close-ended questions that can be answered with a minimal answer (\eg, yes or no~\cite{clark2019boolq}).

Researchers have developed a variety of free-form multiple-choice MRC datasets that contain a significant percentage of questions focusing on the \textbf{implicitly} expressed facts, events, opinions, or emotions in the given text~\cite{richardson2013mctest,lai2017race,ostermann2018semeval,khashabi2018looking,sundream2018}. Generally, we require the integration of our own prior knowledge and the information presented in the given text to answer these questions, posing new challenges for MRC systems. However, until recently, progress in the development of techniques for addressing this kind of MRC task for Chinese has lagged behind their English counterparts. A primary reason is that most previous work focuses on constructing MRC datasets for Chinese in which most answers are either spans~\cite{cui2016consensus,li2016dataset,cui2018dataset,shao2018drcd} or abstractive texts~\cite{he2017dureader} merely based on the information \textbf{explicitly} expressed in the provided text.

With a goal of developing similarly challenging, but free-form multiple-choice datasets, and promoting the development of MRC techniques for Chinese, we introduce the first free-form multiple-\textbf{C}hoice \textbf{C}hinese machine reading \textbf{C}omprehension dataset ({\dn}) that not only contains multiple types of questions but also requires both the information in the given document \textbf{and} prior knowledge to answer questions. In particular, for assessing model generalizability across different domains, {\dn} includes a dialogue-based task {\dnd} in which the given document is a \textbf{d}ialogue, and a mixed-genre task {\dnm} in which the given document is a \textbf{m}ixed-genre text that is relatively formally written. All problems are collected from real-world Chinese-as-a-second-language examinations carefully designed by experts to test the reading comprehension abilities of language learners of Chinese.

We perform an in-depth analysis of what kinds of prior knowledge are needed for answering questions correctly in {\dn} and two representative free-form multiple-choice MRC datasets for English~\cite{lai2017race,sundream2018}, and to what extent. We find that solving these general-domain problems requires linguistic knowledge, domain-specific knowledge, and general world knowledge, the latter of which can be further broken down into eight types such as arithmetic, connotation, cause-effect, and implication. These free-form MRC datasets exhibit similar characteristics in that (i) they contain a high percentage (\eg, $86.8\%$ in {\dn}) of questions requiring knowledge gained from the accompanying document as well as at least one type of prior knowledge and (ii) regardless of language, dialogue-based MRC tasks tend to require more general world knowledge and less linguistic knowledge compared to tasks accompanied with relatively formally written texts. Specifically, compared to existing MRC datasets for Chinese~\cite{he2017dureader,cui2018span}, {\dn} requires more general world knowledge ($57.3\%$ of questions) to arrive at the correct answer options. %

We implement rule-based and popular neural approaches to the MRC task and find that there is still a significant performance gap between the best-performing model ($68.5\%$) and human readers ($96.0\%$), especially on problems that require prior knowledge. We find that the existence of wrong answer options that highly superficially match the given text plays a critical role in increasing the difficulty level of questions and the demand for prior knowledge. Furthermore, additionally introducing $94$k training instances based on translated free-form multiple-choice datasets for English can only lead to a $4.6\%$ improvement in accuracy, still far from closing the gap to human performance. Our hope is that {\dn} can serve as a platform for researchers interested in studying how to leverage different types of prior knowledge for in-depth text comprehension and facilitate future work on crosslingual and multilingual machine reading comprehension.
\section{Related Work}
\label{related}

\begin{table*}[ht!]
\centering
\scriptsize
\begin{tabular}{llllll}
\toprule
\bf Chinese Task               & \bf Document              & \bf Question     &\bf Answer  & \bf Question & \bf English Counterpart\\
               & \bf Genre             & \bf Type     &\bf Type & \bf Size  & \\
\midrule
\bf Question Answering\\
\midrule
QS~\cite{chengtaking}   & N/A           & free-form          & multiple-choice            & $0.6$K  & ARC~\cite{clark2016combining} \\      
MCQA~\cite{guo2017ijcnlp}    & N/A       & free-form           & multiple-choice          & $14.4$K  & ARC~\cite{clark2016combining} \\
MedQA~\cite{zhang2018medical}   & N/A    & free-form    & multiple-choice                 & $235.2$K & ARC~\cite{clark2016combining} \\
GeoSQA~\cite{huang2019geosqa}   & N/A    & free-form    & multiple-choice                 & $4.1$K  & DD~\cite{lally2017watsonpaths} \\
\midrule
\bf Machine Reading Comprehension\\
\midrule
PD~\cite{cui2016consensus}     & news          & cloze         & extractive        &  $876.7$K        & CNN/Daily~\cite{hermann2015teaching} \\
CFT~\cite{cui2016consensus}    & books         & cloze          & extractive       &   $3.6$K        & CBT~\cite{hill2015goldilocks} \\
CMRC 2018~\cite{cui2018span}   & Wiki         & free-form       & extractive       &  $19.1$K              & SQuAD~\cite{rajpurkar2016squad} \\
DuReader~\cite{he2017dureader} & web          & free-form          & abstractive   &  $\approx 200$K   &  MS MARCO~\cite{nguyen2016ms}  \\ %
ChID~\cite{zheng2019chid}             & mixed-genre    & cloze  & multiple-choice                &  $728.7$K & CLOTH~\cite{xie2018large} \\

\bf {\dnm} (this work)                     & mixed-genre    &  free-form             & multiple-choice   &  $10.0$K      & RACE~\cite{lai2017race}   \\
\bf {\dnd} (this work)                    & dialogue        &  free-form            & multiple-choice    &  $9.6$K     & DREAM~\cite{sundream2018}   \\
\bottomrule
\end{tabular}
\caption{Comparison of {\dn} and representative Chinese question answering and machine reading comprehension tasks. We list only one English counterpart for each Chinese dataset.} %

\label{tab:related}
\end{table*}

Traditionally, MRC tasks have been designed to be \textbf{text-dependent}~\cite{richardson2013mctest,hermann2015teaching}: they focus on evaluating comprehension of machine readers based on \textbf{a given text}, typically by requiring a model to answer questions relevant to the text. This is distinguished from many question answering (QA) tasks~\cite{fader2014open,clark2016combining}, in which \textbf{no} ground truth document supporting answers is provided with each question, making them relatively less suitable for isolating improvements to MRC. We will first discuss standard MRC datasets for English, followed by MRC/QA datasets for Chinese.

\label{related:mrc}

\paragraph{English.} 

Much of the early MRC work focuses on designing questions whose answers are spans from the given documents~\cite{hermann2015teaching,hill2015goldilocks,bajgar2016embracing,rajpurkar2016squad,trischler2017newsqa,triviaQA}. As a question and its answer are usually in the same sentence, state-of-the-art methods~\cite{bert2018} have outperformed human performance on many such tasks. To increase task difficulty, researchers have explored a number of options including adding unanswerable~\cite{trischler2017newsqa,rajpurkar2018squad} or conversational~\cite{choi2018quac,reddy2018coqa} questions that might require reasoning~\cite{zhang2018record}, and designing abstractive answers~\cite{nguyen2016ms,kovcisky2018narrativeqa,proparNaacl2018} or (question, answer) pairs that involve cross-sentence or cross-document content~\cite{welbl2018constructing,yang2018hotpotqa}. In general, most questions concern the facts that are explicitly expressed in the text, making these tasks possible to measure the level of fundamental reading skills of machine readers.

Another research line has studied MRC tasks, usually in a free-form multiple-choice form, containing a significant percentage of questions that focus on the understanding of the implicitly expressed facts, events, opinions, or emotions in the given text~\cite{richardson2013mctest,mostafazadeh2016corpus,khashabi2018looking,lai2017race,sundream2018}. Therefore, these benchmarks may allow a relatively comprehensive evaluation of different reading skills and require a machine reader to integrate prior knowledge with information presented in a text. In particular, real-world language exams are ideal sources for constructing this kind of MRC datasets as they are designed with a similar goal of measuring different reading comprehension abilities of human language learners primarily based on a given text. Representative datasets in this category include RACE~\cite{lai2017race} and DREAM~\cite{sundream2018}, both collected from English-as-a-foreign-language exams designed for Chinese learners of English. {\dnm} and {\dnd} can be regarded as a Chinese counterpart of RACE and DREAM, respectively, and we will discuss their similarities in detail in Section~\ref{data:knowledge}.  %

\paragraph{Chinese.} Extractive MRC datasets for Chinese~\cite{cui2016consensus,li2016dataset,cui2018span,cui2018dataset,shao2018drcd} have also been constructed --- using web documents, news reports, books, and Wikipedia articles as source documents --- and for which all answers are spans or sentences from the given documents.~\newcite{zheng2019chid} propose a cloze-style multiple-choice MRC dataset by replacing idioms in a document with blank symbols, and the task is to predict the correct idiom from candidate idioms that are similar in meanings. The abstractive dataset DuReader~\cite{he2017dureader} contains questions collected from query logs, free-form answers, and a small set of relevant texts retrieved from web pages per question. In contrast, {\dn} is the first free-form multiple-choice Chinese MRC dataset that contains different types of questions and requires rich prior knowledge especially general world knowledge for a better understanding of the given text. Furthermore, $48.4\%$ of problems require dialogue understanding, which has not been studied yet in existing Chinese MRC tasks.

Similarly, questions in many existing multiple-choice QA datasets for Chinese~\cite{chengtaking,guo2017ijcnlp,guo2017effective,zhang2018one,zhang2018medical,hao2018exploiting,huang2019geosqa} are also free-form and collected from exams. However, most of the pre-existing QA tasks for Chinese are designed to test the acquisition and exploitation of domain-specific (\eg, history, medical, and geography) knowledge rather than general reading comprehension, and the performance of QA systems is partially dependent on the performance of information retrieval or the relevance of external resource (\eg, corpora or knowledge bases). We compare {\dn} with relevant MRC/QA datasets for Chinese and English in Table~\ref{tab:related}.  

\section{Data}
\label{data}

In this section, we describe the construction of {\dn} (Section~\ref{data:collection}). We also analyze the data (Section~\ref{data:analysis}) and the types of prior knowledge needed for the MRC tasks (Section~\ref{data:knowledge}).

\subsection{Collection Methodology and Task Definitions}
\label{data:collection}

\hhide{
顿时，教室里爆发出了一阵善意的笑声，随即一阵鼓励的掌声响起。得知这件事之后，胡适对沈从文大加赞赏，认为他非常成功。& Immediately, a peal of friendly laughter filled the classroom. Presently, a round of encouraging applause was given to him. Hearing this episode, Hu heaped praise upon Shen, thinking that he was very successful. \\
有了这次经历，在以后的课堂上，沈从文都会告诫自己不要紧张，渐渐地，他开始在课堂上变得从容起来。& Because of this experience, Shen always reminded himself of not being nervous in his class for years afterwards. Gradually, he began to give his lecture at leisure in class. \\}

\begin{table*}[ht!]
\centering
\scriptsize
\begin{tabular}{p{6cm}p{9cm}}
\toprule
1928年，经徐志摩介绍，时任中国公学校长的胡适聘用了沈从文做讲师，主讲大学一年级的现代文学选修课。& In 1928, recommended by Hsu Chih-Mo, Hu Shih, who was the president of the previous National University of China, employed Shen Ts'ung-wen as a lecturer of the university in charge of teaching the optional course of modern literature. \\
当时，沈从文已经在文坛上崭露头角，在社会上也小有名气，因此还未到上课时间，教室里就坐满了学生。上课时间到了，沈从文走进教室，看见下面黑压压一片，心里陡然一惊，脑子里变得一片空白，连准备了无数遍的第一句话都堵在嗓子里说不出来了。& $~~$ At that time, Shen already made himself conspicuous in the literary world and was a little famous in society. For this sake, even before the beginning of class, the classroom was crowded with students. Upon the arrival of class, Shen went into the classroom. Seeing \textbf{a dense crowd} of students sitting beneath the platform, Shen was suddenly startled and his mind went blank. He was even unable to utter the first sentence he had rehearsed repeatedly. \\
他呆呆地站在那里，面色尴尬至极，双手拧来拧去无处可放。上课前他自以为成竹在胸，所以就没带教案和教材。整整 10 分钟，教室里鸦雀无声，所有的学生都好奇地等着这位新来的老师开口。沈从文深吸了一口气， 慢慢平静了下来，原先准备好的东西也重新在脑子里聚拢，然后他开始讲课了。不过由于他依然很紧张，原本预计一小时的授课内容，竟然用了不到 15 分钟就讲完了。& $~~~$ He stood there motionlessly, extremely embarrassed. He wrung his hands without knowing where to put them. Before class, he believed that he had a ready plan to meet the situation so he did not bring his teaching plan and textbook. For up to 10 minutes, the classroom was in perfect silence. All the students were curiously waiting for the new teacher to open his mouth. Breathing deeply, he gradually calmed down. Thereupon, the materials he had previously prepared gathered in his mind for the second time. Then he began his lecture. Nevertheless, since he was still nervous, it took him less than 15 minutes to finish the teaching contents he had planned to complete in an hour. \\
接下来怎么办？他再次陷入了窘境。无奈之下，他只好拿起粉笔在黑板上写道：我第一次上课，见你们人多，怕了。& $~~$ What should he do next? He was again caught in embarrassment. He had no choice but to pick up a piece of chalk before writing several words on the blackboard: This is the first time I have given a lecture. In the presence of a crowd of people, I feel terrified. \\
顿时，教室里爆发出了一阵善意的笑声，随即一阵鼓励的掌声响起。得知这件事之后，胡适对沈从文大加赞赏，认为他非常成功。有了这次经历，在以后的课堂上，沈从文都会告诫自己不要紧张，渐渐地，他开始在课堂上变得从容起来。 & $~~~$ Immediately, a peal of friendly laughter filled the classroom. Presently, a round of encouraging applause was given to him. Hearing this episode, Hu heaped praise upon Shen, thinking that he was very successful. Because of this experience, Shen always reminded himself of not being nervous in his class for years afterwards. Gradually, he began to give his lecture at leisure in class. \\
\midrule
\textbf{Q1}\hspace{2mm}第2段中， “黑压压一片”指的是：& \textbf{Q1}\hspace{2mm}In paragraph 2, \emph{``a dense crowd''} refers to\\
A.\hspace{3mm}教室很暗 & A.\hspace{3mm}the light in the classroom was dim. \\
B.\hspace{3mm}听课的人多 $\star$ & B.\hspace{3mm}the number of students attending his lecture was large. $\star$\\
C.\hspace{3mm}房间里很吵 & C.\hspace{3mm}the room was noisy.\\
D.\hspace{3mm}学生们发言很积极 & D.\hspace{3mm}the students were active in voicing their opinions.\\

\textbf{Q2}\hspace{2mm}沈从文没拿教材，是因为他觉得：& \textbf{Q2}\hspace{2mm}Shen did not bring the textbook because he felt that\\
A.\hspace{3mm}讲课内容不多 & A.\hspace{3mm}the teaching contents were not many. \\
B.\hspace{3mm}自己准备得很充分 $\star$ & B.\hspace{3mm}his preparation was sufficient. $\star$\\ 
C.\hspace{3mm}这样可以减轻压力 & C.\hspace{3mm}his mental pressure could be reduced in this way.\\
D.\hspace{3mm}教材会限制自己的发挥 & D.\hspace{3mm}the textbook was likely to restrict his ability to give a lecture.\\

\textbf{Q3}\hspace{2mm}看见沈从文写的那句话，学生们：& \textbf{Q3}\hspace{2mm}Seeing the sentence written by Shen, the students\\
A.\hspace{3mm}急忙安慰他 & A.\hspace{3mm}hurriedly consoled him. \\
B.\hspace{3mm}在心里埋怨他 & B.\hspace{3mm}blamed him in mind. \\
C.\hspace{3mm}受到了极大的鼓舞 & C.\hspace{3mm}were greatly encouraged.\\
D.\hspace{3mm}表示理解并鼓励了他 $\star$ & D.\hspace{3mm}expressed their understanding and encouraged him. $\star$\\

\textbf{Q4}\hspace{2mm}上文主要谈的是：& \textbf{Q4}\hspace{2mm}The passage above is mainly about\\
A.\hspace{3mm}中国教育制度的发展 & A.\hspace{3mm}the development of the Chinese educational system. \\
B.\hspace{3mm}紧张时应如何调整自己 & B.\hspace{3mm}how to make self-adjustment if one is nervous. \\
C.\hspace{3mm}沈从文第一次讲课时的情景 $\star$ & C.\hspace{3mm}the situation where Shen gave his lecture for the first time. $\star$\\
D.\hspace{3mm}沈从文如何从作家转变为教师的 & D.\hspace{3mm}how Shen turned into a teacher from a writer.\\

\bottomrule
\end{tabular}

\caption{A {\dn}-Mixed ({\dnm}) problem (left) and its English translation (right) ($\star$: the correct option).} %
\label{tab:sample1a-parallel}
\end{table*}

\begin{table}[ht!]
\centering
\scriptsize
\begin{tabular}{p{0.2cm}p{6.5cm}}
\toprule
\textbf{F}: & How is it going? Have you bought your ticket? \\
\textbf{M}: & There are so many people at the railway station. I have waited in line all day long. However, when my turn comes, they say that there is no ticket left unless the Spring Festival is over. \\
\textbf{F}: & It doesn't matter. It is all the same for you to come back after the Spring Festival is over. \\
\textbf{M}: & But according to our company's regulation, I must go to the office on the 6th day of the first lunar month. I'm afraid I have no time to go back after the Spring Festival, so could you and my dad come to Shanghai for the coming Spring Festival?\\
\textbf{F}: & I am too old to endure the travel.\\
\textbf{M}: & It is not difficult at all. After I help you buy the tickets, you can come here directly.  \\
\midrule
\multicolumn{2}{l}{\textbf{Q1}\hspace{2mm}What is the relationship between the speakers?}\\
\multicolumn{2}{l}{A.\hspace{3mm}father and daughter.} \\
\multicolumn{2}{l}{B.\hspace{3mm}mother and son. $\star$}\\
\multicolumn{2}{l}{C.\hspace{3mm}classmates.}\\
\multicolumn{2}{l}{D.\hspace{3mm}colleagues.}\\

\multicolumn{2}{l}{\textbf{Q2}\hspace{2mm}What difficulty has the male met?}\\
\multicolumn{2}{l}{A.\hspace{3mm}his company does not have a vacation.} \\
\multicolumn{2}{l}{B.\hspace{3mm}things are expensive during the Spring Festival.} \\ 
\multicolumn{2}{l}{C.\hspace{3mm}he has not bought his ticket. $\star$}\\
\multicolumn{2}{l}{D.\hspace{3mm}he cannot find the railway station.}\\

\multicolumn{2}{l}{\textbf{Q3}\hspace{2mm}What suggestion does the male put forth?} \\
\multicolumn{2}{l}{A.\hspace{3mm}he invites the female to come to Shanghai. $\star$} \\
\multicolumn{2}{l}{B.\hspace{3mm}he is going to wait in line the next day.} \\
\multicolumn{2}{l}{C.\hspace{3mm}he wants to go to the company as soon as possible.} \\
\multicolumn{2}{l}{D.\hspace{3mm}he is going to go home after the Spring Festival is over.} \\

\bottomrule
\end{tabular}

\caption{English translation of a sample problem from {\dn}-Dialogue ({\dnd}) ($\star$: the correct option).
}
\label{tab:sample1b-en}
\end{table}

\begin{table*}[ht!]
\centering
\scriptsize
\begin{tabular}{llll}
\toprule
 \textbf{Metric} & \textbf{{\dnm}} & \textbf{{\dnd}} & \textbf{{\dn}} \\
\midrule 
Min./Avg./Max. \# of options per question & 2 / 3.7 / 4 & 3 / 3.8 / 4  & 2 / 3.8 / 4 \\
\# of correct options per question & 1 &  1   & 1\\
Min./Avg./Max. \# of questions per document & 1 / 1.9 / 6 & 1 / 1.2 / 6 & 1 / 1.5 / 6 \\
Avg./Max. option length (in characters) & 6.5 / 45 & 4.4 / 31  & 5.5 / 45 \\
Avg./Max. question length (in characters) & 13.5 / 57 & 10.9 / 34  & 12.2 / 57 \\
Avg./Max. document length (in characters) & 180.2 / 1,274 & 76.3 / 1,540 &  116.9 / 1,540 \\
character vocabulary size & 4,120 & 2,922  & 4,193 \\
non-extractive correct option (\%) & 81.9 & 78.9  & 80.4 \\
\midrule
    \multicolumn{4}{l}{\textbf{\# of documents / \# of questions}} \\
~~Training & 3,138 / 6,013 & 4,885 / 5,856  & 8,023 / 11,869\\
~~Development & 1,046 / 1,991 & 1,628 / 1,825  & 2,674 / 3,816\\
~~Test & 1,045 / 2,002 & 1,627 / 1,890  & 2,672 / 3,892\\
~~All & 5,229 / 10,006 & 8,140 / 9,571  & 13,369 / 19,577\\
\bottomrule

\end{tabular}
\caption{The overall statistics of {\dn}. {\dn} $=$ {\dnm} $\cup$ {\dnd}.}
\label{tab:data:stat}
\end{table*}

We collect the general-domain problems from Hanyu Shuiping Kaoshi (HSK) and Minzu Hanyu Kaoshi (MHK), which are designed for evaluating the Chinese listening and reading comprehension ability of second-language learners such as international students, overseas Chinese, and ethnic minorities. %
We include problems from both real and practice exams; all are freely accessible online for public usage.

Each problem consists of a document and a series of questions. Each question is associated with several answer options, and \textsc{exactly one} of them is correct. The goal is to select the correct option. According to the document type, we divide these problems into two subtasks:  {\dn}-Dialogue ({\dnd}), in which a dialogue serves as the document, and {\dn}-Mixed ({\dnm}), in which the given non-dialogue document is of mixed genre, such as a story, a news report, a monologue, or an advertisement. %
We show a sample problem for each type in Tables~\ref{tab:sample1a-parallel} and~\ref{tab:sample1b-en}, respectively.

We remove duplicate problems and randomly split the data (13,369 documents and 19,577 questions in total) at the problem level, with $60\%$ training, $20\%$ development, and $20\%$ test.

\subsection{Data Statistics}
\label{data:analysis}

We summarize the overall statistics of {\dn} in Table~\ref{tab:data:stat}. We observe notable differences exist between {\dnm} and {\dnd}. For example, {\dnm}, in which most documents are formally written texts, has a larger vocabulary size compared to that of {\dnd} with documents in spoken language. Similar observations have been made by~\newcite{sundream2018} that the vocabulary size is relatively small in English dialogue-based machine reading comprehension tasks. In addition, the average document length ($180.2$) in {\dnm} is longer than that in {\dnd} ($76.3$). In general, {\dn} may not be suitable for evaluating the comprehension ability of machine readers on lengthy texts as the average length of document {\dn} is relatively short compared to that in datasets such as DuReader~\cite{he2017dureader} ($396.0$) and RACE~\cite{lai2017race} ($321.9$).

\begin{table*}[ht!]
\scriptsize
\centering
\begin{tabular}{lllllll}
\toprule
  & \bf \dnm & \bf \dnd    & \bf \dn  & \bf RACE &  \bf DREAM   & \bf DuReader\\
\midrule
Matching                & 12.0 & 14.3   & 13.2 & 14.7 & 8.7  & 62.0\\
\midrule
Prior knowledge         & 88.0 & 85.7   & 86.8 & 85.3 & 91.3   & 38.0 \\
$~~~~\diamond$ Linguistic              & \textbf{49.0} & 30.7   & 39.8 & 47.3 & 40.0    &  22.0\\
$~~~~\diamond$ Domain-specific         & 0.7  & 1.0    & 0.8 & 0.0 & 0.0  & 16.0 \\
$~~~~\diamond$ General world           & 50.7 & \textbf{64.0}  & 57.3 & 43.3 &  57.3   & 0.0 \\
$~~~~~~~~~~$ Arithmetic   & 3.0  & 4.7    & 3.8  & 3.3 &  1.3   &  0.0\\
$~~~~~~~~~~$ Connotation  & 1.3  & 5.3     & 3.3 & 2.0 & 5.3    &  0.0\\
$~~~~~~~~~~$ Cause-effect & 14.0 & 6.7    & 10.3 & 2.7 & 3.3   & 0.0 \\
$~~~~~~~~~~$ Implication  & 17.7 & 20.3  & 19.0 & 24.0 & 26.7   & 0.0\\
$~~~~~~~~~~$ Part-whole   & 5.0  & 5.0     & 5.0 & 2.7 & 7.3  & 0.0\\
$~~~~~~~~~~$ Precondition & 2.7  & 4.3    & 3.5 & 2.7 & 1.3   & 0.0\\
$~~~~~~~~~~$ Scenario     & 9.6  & \textbf{24.3}   & 17.0  & 7.3 & 21.3   & 0.0\\
$~~~~~~~~~~$ Other        & 3.3  & 0.3    & 1.8   & 2.0 &  0.7   & 0.0\\
\midrule
Single sentence         & 50.7 & 22.7  & 36.7 & 24.0 &  12.0   & 14.6\\
Multiple sentences      & 47.0 & 77.0  & 62.0 & 75.3 &  88.0   & 68.7\\
Independent             & 2.3  & 0.3  & 1.3 & 0.7 &  0.0 & 16.7\\
\midrule
\# of annotated instances  & 300   & 300   & 600 & 150 &  150  & 150\\
\bottomrule
\end{tabular}
\caption{Distribution (\%) of types of required prior knowledge based on a subset of test and development sets of {\dn}, Chinese free-form abstractive dataset DuReader~\cite{he2017dureader}, and English free-form multiple-choice datasets RACE~\cite{lai2017race} and DREAM~\cite{sundream2018}. Answering a question may require more than one type of prior knowledge. }
\label{tab:data:category}
\end{table*}

\subsection{Categories of Prior Knowledge}
\label{data:knowledge}

Previous studies on Chinese machine reading comprehension focus mainly on the linguistic knowledge required~\cite{he2017dureader,cui2018dataset}. We aim instead for a more comprehensive analysis of the types of prior knowledge for answering questions. We carefully analyze a subset of questions randomly sampled from the development and test sets of {\dn} and arrive at the following three kinds of prior knowledge required for answering questions. A question is labeled as \textbf{matching} if it exactly matches or nearly matches (without considering determiners, aspect particles, or conjunctive adverbs~\cite{xia2000part}) a span in the given document; answering questions in this category seldom requires any prior knowledge.

\noindent \textbf{\textsc{Linguistic}}: To answer a given question (\eg, Q $1$-$2$ in Table~\ref{tab:sample1a-parallel} and Q$3$ in Table~\ref{tab:sample1b-en}), we require lexical/syntactic knowledge including but not limited to: idioms, proverbs, negation, antonymy, synonymy, the possible meanings of the word, and syntactic transformations~\cite{nassaji2006relationship}.

\noindent \textbf{\textsc{Domain-Specific}}: This kind of world knowledge consists of, but is not limited to, facts about domain-specific concepts, their definitions and properties, and relations among these concepts~\cite{grishman1983isolating,hansen1994reasoning}.

\noindent \textbf{\textsc{General World}}: It refers to the general knowledge about how the world works, sometimes called commonsense knowledge. We focus on the sort of world knowledge that an encyclopedia would assume readers know \textbf{without being told}~\cite{lenat1985cyc,schubert2002can} instead of the factual knowledge such as properties of famous entities. We further break down general world knowledge into eight subtypes, some of which (marked with $\dag$) are similar to the categories summarized by~\newcite{lobue2011types} for textual entailment recognition.

\begin{itemize}
    \item Arithmetic$^\dag$: This includes numerical computation and analysis (\eg, comparison and unit conversion).
    \item Connotation: Answering questions requires knowledge about implicit and implied sentiment towards something or somebody, emotions, and tone~\cite{edmonds2002near,feng2013connotation,van2018we}. For example, the following conversation: \emph{``F: Ming Yu became a manager when he was so young! That's impressive! M: It is indeed not easy!''} is delivered in a tone for praise.
    \item Cause-effect$^\dag$: The occurrence of event A causes the occurrence of event B. We usually need this kind of knowledge to solve ``why'' questions when a causal explanation is not explicitly expressed in the given document.
    \item Implication: This category refers to the main points, suggestions, opinions, facts, or event predictions that are not expressed explicitly in the text, which cannot be reached by paraphrasing sentences using linguistic knowledge. For example, Q$4$ in Table~\ref{tab:sample1a-parallel} and Q$2$ in Table~\ref{tab:sample1b-en} belong to this category. 
    \item Part-whole: We require knowledge that object A is a part of object B. Relations such as member-of, stuff-of, and component-of between two objects also fall into this category~\cite{winston1987taxonomy,miller1998wordnet}. For example, we require implication mentioned above as well as part-whole knowledge (\ie, \emph{``teacher''} is a kind of job) to summarize the main topic of the following dialogue as \emph{``profession''}: \emph{``F: Many of my classmates become teachers after graduation. M: The best thing about being a teacher is feeling happy every day as you are surrounded by students!''}.
    \item Scenario: We require knowledge about observable behaviors or activities of humans and their corresponding temporal/locational information. We also need knowledge about personal information (\eg, profession, education level, personality, and mental or physical status) of the involved participant and relations between the involved participants, implicitly indicated by the behaviors or activities described in texts. For example, we put Q$3$ in Table~\ref{tab:sample1a-parallel} in this category as \emph{``friendly laughter''} may express \emph{``understanding''}. Q$1$ in Table~\ref{tab:sample1b-en} about the relation between the two speakers also belongs to this category.
    \item Precondition$^\dag$: If had event A not happened, event B would not have happened~\cite{ikuta2014challenges,o2016richer}. Event A is usually mentioned in either the question or the correct answer option(s). For example, \emph{``I went to a supermarket''} is a necessary precondition for \emph{``I was shopping at a supermarket when my friend visited me''}. 
    \item Other: Knowledge that belongs to none of the above subcategories.
\end{itemize}

Two annotators (authors of this paper) annotate the type(s) of required knowledge for each question over $600$ instances. To explore the differences and similarities in the required knowledge types between {\dn} and existing free-form MRC datasets, following the same annotation schema, we also annotate instances from the largest Chinese free-form abstractive MRC dataset DuReader~\cite{he2017dureader} and free-form multiple-choice English MRC dataset RACE~\cite{lai2017race} and DREAM~\cite{sundream2018} that can be regarded as the English counterpart of {\dnm} and {\dnd}, respectively. We also divide questions into one of three types -- single, multiple, or independent -- based on the minimum number of sentences in the document that explicitly or implicitly support the correct answer option. We regard a question as independent if it is context-independent, which usually requires prior vocabulary or domain-specific knowledge. The Cohen's kappa coefficient is $0.62$.

\bigskip

\noindent \textbf{{\dnm} vs. {\dnd}}
As shown in Table~\ref{tab:data:category}, compared to the dialogue-based task ({\dnd}), {\dnm} with non-dialogue texts as documents requires more linguistic knowledge ($49.0\%$ \vs$30.7\%$) yet less general world knowledge ($50.7\%$ \vs$64.0\%$). As many as $24.3\%$ questions in {\dnd} need scenario knowledge perhaps due to that speakers in a dialogue (especially face-to-face) may not explicitly mention information that they assume others already know such as personal information, the relationship between the speakers, and temporal and location information. Interestingly, we observe a similar phenomenon when we compare the English datasets DREAM (dialogue-based) and RACE. Therefore, it is likely that dialogue-based free-form tasks can serve as ideal platforms for studying how to improve language understanding with general world knowledge regardless of language.

\noindent \textbf{{\dn} vs. its English counterparts}
We are also interested in whether answering a specific type of question may require similar types of prior knowledge across languages. For example, {\dnd} and its English counterpart DREAM~\cite{sundream2018} have similar problem formats, document types, and data collection methodologies (from Chinese-as-a-second-language and English-as-a-foreign-language exams, respectively). We notice that the knowledge type distributions of the two datasets are indeed very similar. Therefore, {\dn} may facilitate future cross-lingual MRC studies. %

\noindent \textbf{{\dn} vs. DuReader~~} The $150$ annotated instances of DuReader also exhibit properties similar to those identified in studies of abstractive MRC for English~\cite{nguyen2016ms,kovcisky2018narrativeqa,reddy2018coqa}. Namely, turkers asked to write answers in his/her own words tend instead to write an extractive summary by copying short textual snippets or whole sentences in the given documents; this may explain why models designed for extractive MRC tasks achieve reasonable performance on abstractive tasks. We notice that questions in DuReader seldom require general world knowledge, which is possibly because users seldom ask questions about facts obvious to most people. On the other hand, as many as $16.7\%$ of (question, answer) pairs in DuReader cannot be supported by the given text (vs. $1.3\%$ in {\dn}); in most cases, they require prior knowledge about a particular domain (\eg, \emph{``On which website can I watch The Glory of Tang Dynasty?''} and \emph{``How to start a clothing store?''}). In comparison, a larger fraction of {\dn} requires linguistic knowledge or general world knowledge.

\begin{table*}[ht!]
\centering
\footnotesize
\begin{tabular}{lcccccc}
\toprule
\multirow{2}{*}{\textbf{Method}} & \multicolumn{2}{c}{\textbf{{\dnm}}} & \multicolumn{2}{c}{\textbf{{\dnd}}}  & \multicolumn{2}{c}{\textbf{\dn}}  \\
& \textbf{Dev} & \textbf{Test} & \textbf{Dev} & \textbf{Test} & \textbf{Dev} & \textbf{Test} \\
\midrule
Random                        & 27.8 & 27.8 & 26.4 & 26.6  & 27.1 & 27.2 \\
Distance-Based Sliding Window~\cite{richardson2013mctest} & 47.9 & 45.8 & 39.6 & 40.4  & 43.8 & 43.1 \\
Co-Matching~\cite{wang2018co}                   & 47.0 & 48.2 & 55.5 & 51.4 & 51.0 & 49.8 \\
$\text{BERT}$~\cite{bert2018}       & 65.6 & 64.6  & 65.9 & 64.4 & 65.7 & 64.5 \\ %
$\text{ERNIE}$~\cite{sun2019ernie}  & 63.7 & 63.6 & 67.3 & 64.6 & 65.5 & 64.1 \\ %
$\text{BERT-wwm}$~\cite{chinese-bert-wwm}     & 66.1 & 64.0 & 64.8 & 65.0 & 65.5 & 64.5 \\
$\text{BERT-wwm-ext}$~\cite{chinese-bert-wwm} & 67.9 & 68.0 &  67.7 & 68.9 & 67.8 & 68.5 \\ %
\midrule
Human Performance$^\ast$ & 96.0 & 93.3 & 98.0 & 98.7 & 97.0 & 96.0 \\
\bottomrule                  
\end{tabular}
\caption{Performance of baseline in accuracy (\%) on the {\dn} dataset ($\ast$: based on the annotated subset of test and development sets of {\dn}).}
\label{tab:eval:baseline} 
\end{table*}

\section{Approaches}

\label{methods}

We implement a classical rule-based method and recent state-of-the-art neural models.

\subsection{Distance-Based Sliding Window}

We implement Distance-based Sliding Window~\cite{richardson2013mctest}, a rule-based method that chooses the answer option by taking into account (1) lexical similarity between a \emph{statement} (\ie, a question and an answer option) and the given document with a fixed window size and (2) the minimum number of tokens between occurrences of the question and occurrences of an answer option in the document. This method assumes that a statement is more likely to be correct if there is a shorter distance between tokens within a statement, and more informative tokens in the statement appear in the document.

\subsection{Co-Matching}

We employ Co-Matching~\cite{wang2018co}, a Bi-LSTM-based model for multiple-choice MRC tasks for English. It explicitly treats a question and one of its associated answer options as two sequences and jointly models whether or not the given document matches them. We modify the pre-processing step and adapt this model to MRC tasks for Chinese (Section~\ref{sec:experimental-setting}).

\subsection{Fine-Tuning Pre-Trained Language Models}
\label{method:model}
We also apply the framework of fine-tuning a pre-trained language model on machine reading comprehension tasks~\cite{radfordimproving}. We consider the following four pre-trained language models for Chinese:  Chinese BERT-Base (denoted as $\text{BERT}$)~\cite{bert2018},  Chinese ERNIE-Base (denoted as $\text{ERNIE}$)~\cite{sun2019ernie}, and Chinese BERT-Base with whole word masking during pre-training (denoted as $\text{BERT-wwm}$)~\cite{chinese-bert-wwm} and its enhanced version pre-trained over larger corpora (denoted as $\text{BERT-wwm-ext}$). These models have the same number of layers, hidden units, and attention heads.

Given document $d$, question $q$, and answer option $o_i$, we construct the input sequence by concatenating \texttt{[CLS]}, tokens in $d$, \texttt{[SEP]}, tokens in $q$, \texttt{[SEP]}, tokens in $o_i$, and \texttt{[SEP]}, where \texttt{[CLS]} and \texttt{[SEP]} are the classifier token and sentence separator in a pre-trained language model, respectively. We add an embedding vector $\bm{t_1}$ to each token before the first \texttt{[SEP]} (inclusive) and an embedding vector $\bm{t_2}$ to every other token, where $\bm{t_1}$ and $\bm{t_2}$ are learned during language model pre-training for discriminating sequences. We denote the final hidden state for the first token in the input sequence as $S_i\in \mathbb{R}^{1 \times H}$, where $H$ is the hidden size. We introduce a classification layer $W \in \mathbb{R}^{1 \times H}$ and obtain the unnormalized log probability $P_i\in \mathbb{R}$ of $o_i$ being correct by $P_i=S_iW^T$. We obtain the final prediction for $q$ by applying a softmax layer over the unnormalized log probabilities of all options associated with $q$.
\section{Experiment}
\label{experiments}

 \begin{table*}[ht!]
 \footnotesize
 \centering
 \begin{tabular}{lcccc}
 \toprule
   & \bf Co-Matching & \bf $\text{BERT}$ & \bf $\text{BERT-wwm-ext}$ &  \bf Human\\
   & \textbf{\dnm | \dnd} &  \textbf{\dnm | \dnd} &  \textbf{\dnm | \dnd} & \textbf{\dnm | \dnd}  \\
 \midrule
 Matching            & 54.6 | 70.4  &  81.8 | 81.5  & 100.0 | 85.2~~ & 100.0 | 100.0 \\
 Prior knowledge        & 47.5 | 51.2  & 64.0 | 64.2  & 62.6 | 68.3 & 95.7 | 97.6 \\
 $\diamond$ Linguistic  & 49.4 | 49.0 &  67.1 | 62.8  & 61.2 | 68.6  & ~~97.7 | 100.0\\
 $\diamond$ Domain-specific$^\ast$ & ~~~~--~ | 66.7   & ~~--~  | 0.0 &  ~~--~ | 0.0  & ~~~~~~--~ | 100.0 \\
 $\diamond$ General world   & 46.5 | 53.8  & 57.7 | 66.3  & 64.8  | 70.0 & 93.0 | 96.3 \\
 $~~~~~~$ Arithmetic$^\ast$ & 50.0 | 60.0  & ~~0.0 | 80.0 & 50.0 | 60.0 & 100.0 | 100.0 \\
 $~~~~~~$ Connotation$^\ast$ & ~~0.0 | 50.0 & ~~0.0 | 62.5 & ~~0.0 | 62.5  & 100.0 | 100.0\\
 $~~~~~~$ Cause-effect & 47.6 | 55.6  & 57.1 | 55.6  & 66.7 | 66.7 & ~~95.2 | 100.0 \\
 $~~~~~~$ Implication & 46.7 | 45.5  & 70.0 | 50.0 & 70.0 | 54.6 & 86.7 | 95.5 \\
 $~~~~~~$ Part-whole & 60.0 | 50.0 & 40.0 | 50.0 & 40.0 | 50.0  & 100.0 | 83.3~~ \\
 $~~~~~~$ Precondition$^\ast$ & 66.7 | 50.0  & 66.7 | 25.0 & 66.7 | 75.0  & 100.0 | 100.0 \\
 $~~~~~~$ Scenario & 40.0  | 61.3 & 40.0 | 80.7 & 60.0 | 83.9 & 100.0 | 96.8~~ \\
 $~~~~~~$ Other$^\ast$ & ~~--~ | 0.0 & ~~--~ | 0.0 & ~~--~ | 0.0 &  ~~~~~~--~ | 100.0 \\
 \midrule
Single sentence         & 50.0 | 64.7  &  72.4 | 76.5 & 71.1 | 82.4  & 97.4 | 97.1 \\
Multiple sentences      & 47.2 | 51.7 & 58.3 | 64.7  & 61.1 | 68.1 & 94.4 | 98.3 \\
Independent$^\ast$     & 0.0 | ~--~~ & 50.0 | ~--~~~~ & 0.0 | ~--~~   & 100.0 | ~--~~~~~~ \\
\bottomrule
\end{tabular}
\caption{Performance comparison in accuracy (\%) by categories based on a subset of development sets of {\dn} ($\ast$: $\leq$ 10 annotated instances fall into that category).}
\label{tab:data:gap}
\end{table*}

\begin{figure*}%
    \centering
    \subfloat[Performance comparison based on different largest distractor plausibility.]{{\includegraphics[width=0.4\textwidth]{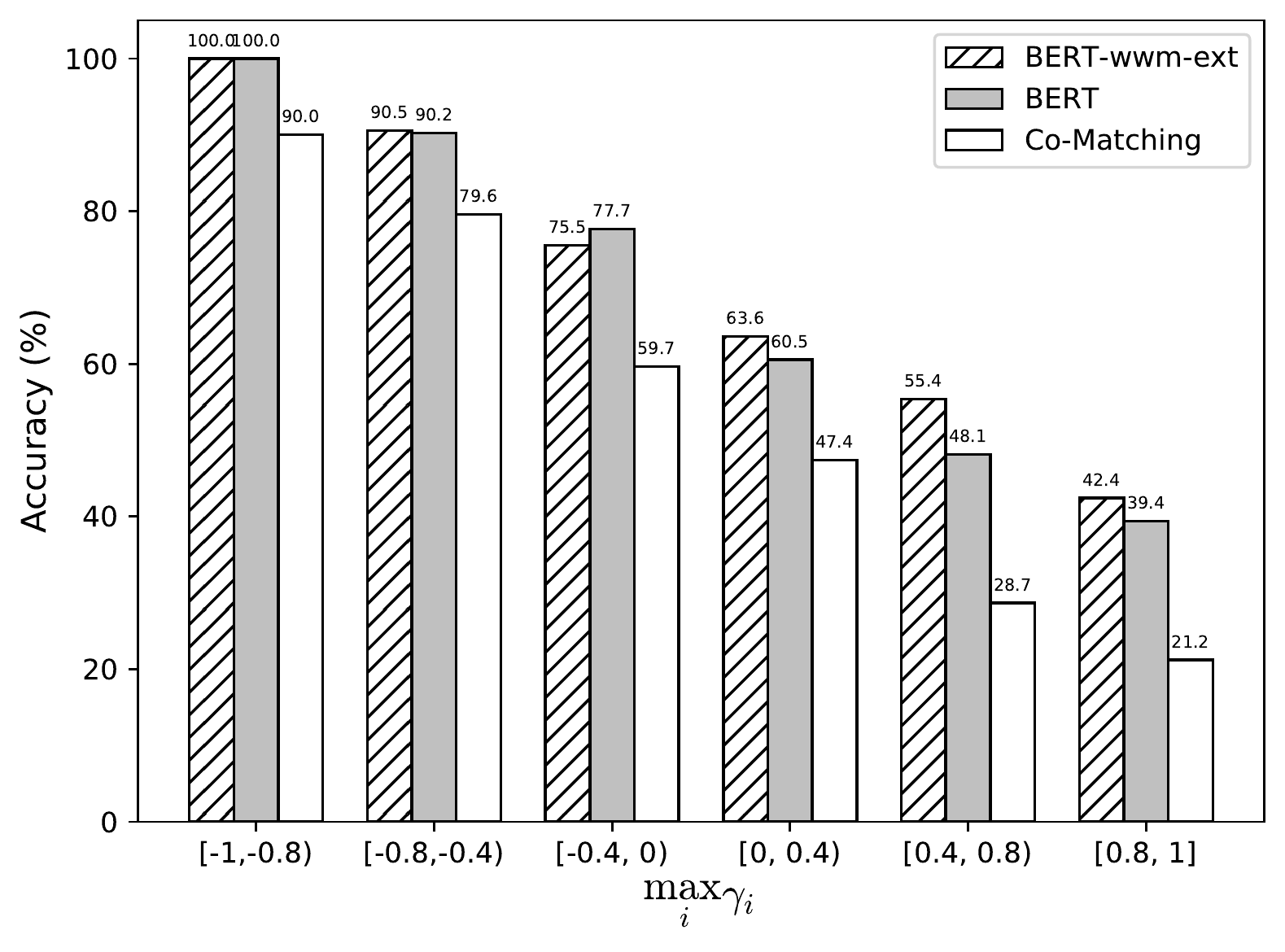} }}%
    \qquad
    \subfloat[Correlation between largest distractor plausibility and the need for prior knowledge.]{{\includegraphics[width=0.4\textwidth]{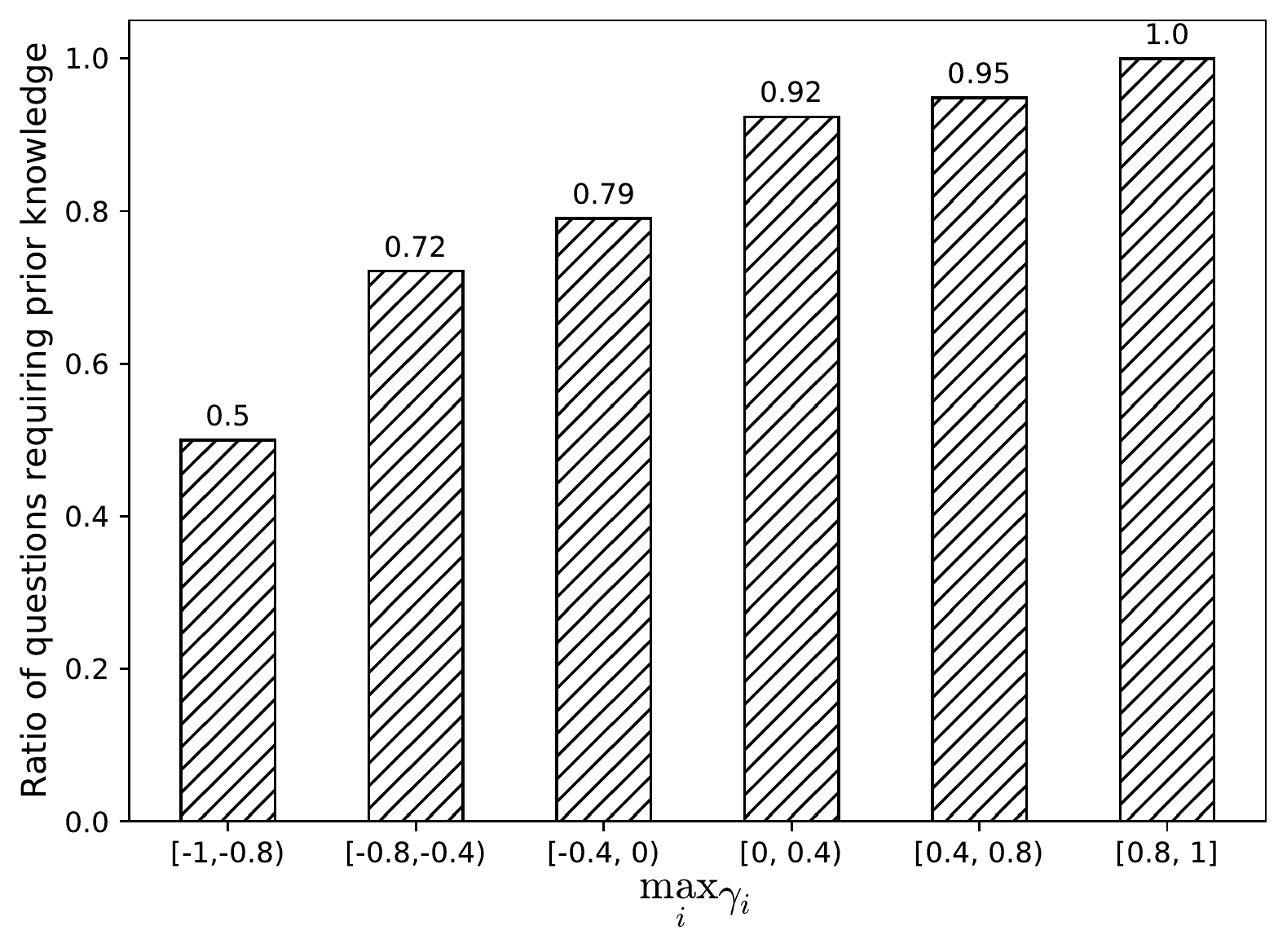} }}%
    \caption{Analysis of distractor plausibility.}
    \label{fig:eval:distractorplaus}%
\end{figure*}

\subsection{Experimental Settings}
\label{sec:experimental-setting}

We use {\dnm} and {\dnd} together to train a neural model and perform testing on them separately, following the default setting on RACE that also contains two subsets~\cite{lai2017race}. We run every experiment five times with different random seeds and report the best development set performance and its corresponding test set performance.

\paragraph{Distance-Based Sliding Window.} We simply treat each character as a token. We do not employ Chinese word segmentation as it results in drops in performance based on our experiment. 

\paragraph{Co-Matching.} We replace the English tokenizer with a Chinese word segmenter in HanLP.\footnote{\url{https://github.com/hankcs/HanLP}.}
We use the $300$-dimensional Chinese word embeddings released by~\newcite{P18-2023}.

\paragraph{Fine-Tuning Pre-Trained Language Models.} We set the learning rate, batch size, and maximal sequence length to $2\times 10^{-5}$, $24$, and $512$, respectively. We truncate the longest sequence among $d$, $q$, and $o_i$ (Section~\ref{method:model}) when an input sequence exceeds the length limit $512$. For all experiments, we fine-tune a model on {\dn} for eight epochs. We keep the default values for the other hyperparameters~\cite{bert2018}. %

\subsection{Baseline Results } 
\label{eval:overview}

As shown in Table~\ref{tab:eval:baseline}, methods based on pre-trained language models (BERT, ERNIE, BERT-wwm, and BERT-wwm-ext) outperform the Distance-based Sliding Window approach and Bi-LSTM-based Co-Matching by a large margin. BERT-wwm-ext performs better on {\dn} compared to other three pre-trained language models, though there still exists a large gap  ($27.5\%$) between this method and human performance ($96.0\%$). 

We also report the performance of Co-Matching, BERT, BERT-wwm-ext, and human on different question categories based on the annotated development sets (Table~\ref{tab:data:gap}), which consist of $150$ questions in {\dnm} and $150$ questions in {\dnd}. These models generally perform worse on questions that require prior knowledge or reasoning over multiple sentences than questions that can be answered by surface matching or only need the information from a single sentence (Section~\ref{data:knowledge}).

\begin{figure*}%
    \centering
    \subfloat[The need for linguistic knowledge.]{{\includegraphics[width=0.4\textwidth]{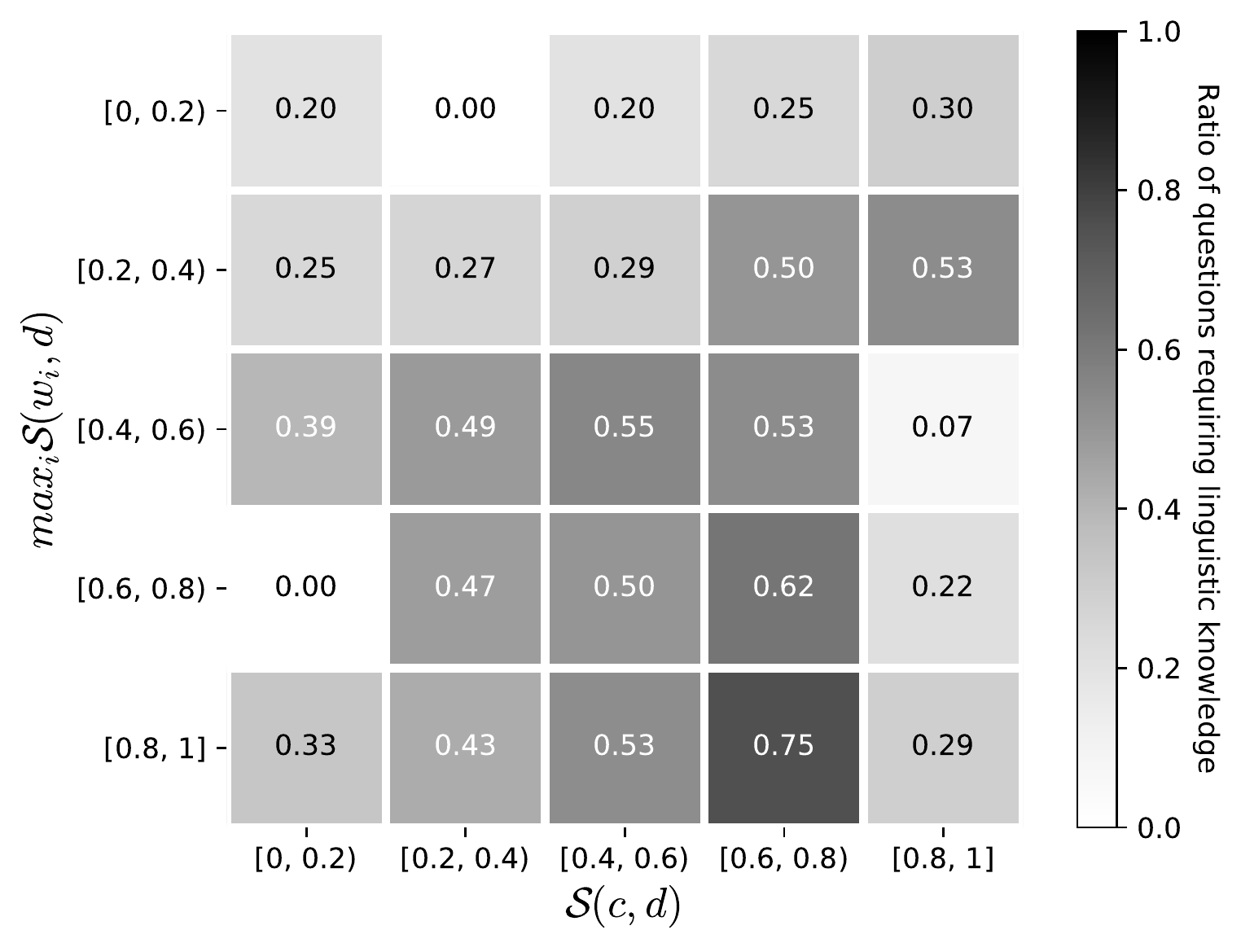} }}%
    \qquad
    \subfloat[The need for general world knowledge.]{{\includegraphics[width=0.4\textwidth]{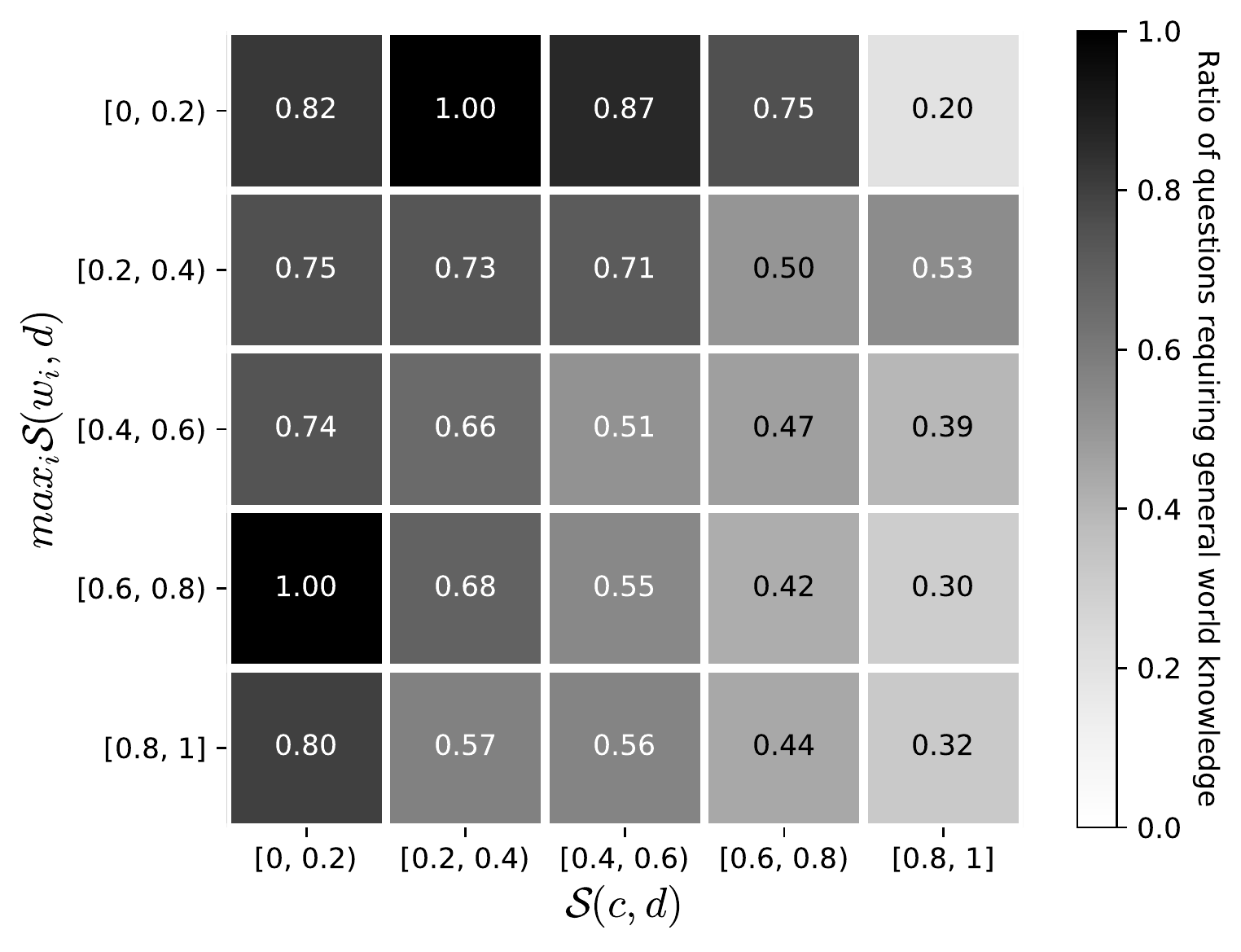} }}%
    \caption{The need for two major types of prior knowledge when answering questions of different $\max_i\mathcal{S}(w_i,d)$ and $\mathcal{S}(c,d)$.}
    \label{fig:eval:distractorsim}%
\end{figure*}

\subsection{Discussions on Distractor Plausibility}

We look into incorrect predictions of Co-Matching, BERT, and BERT-wwm-ext on the development set. We observe that the existence of \textbf{\emph{plausible distractors}} may play a critical role in raising the difficulty level of questions for models. We regard a \textbf{\emph{distractor}} (\ie, wrong answer option) as plausible if it, compared with the correct answer option, is more superficially similar to the given document. Two typical cases include (1) the information in the distractor is accurate based on the document but does not (fully) answer the question, and (2) the distractor distorts, oversimplifies, exaggerates, or misinterprets the information in the document.

Given document $d$, the correct answer option $c$, and wrong answer options $\{w_1,w_2,\ldots , w_i, \ldots, w_n\}$ associated with a certain question, we measure the \textbf{\emph{distractor plausibility}} of distractor $w_i$ by:

\begin{equation}
\gamma_i = \mathcal{S}(w_i,d)-\mathcal{S}(c,d) \label{eq:dd}
\end{equation}
where $\mathcal{S}(x,y)$ is a normalized similarity score between $0$ and $1$ that measures the edit distance to change $x$ into a substring of $y$ using single-character edits (insertions, deletions or substitutions). Particularly, if $x$ is a substring of $y$, $\mathcal{S}(x,y)=1$; if $x$ shares no character with $y$, $\mathcal{S}(x,y)=0$. By definition, $\mathcal{S}(w_i,d)$ in Equation~(\ref{eq:dd}) measures the lexical similarity between distractor $w_i$ and $d$; $\mathcal{S}(c,d)$ measures the similarity between the correct answer option $c$ and $d$.

To quantitatively investigate the impact of the existence of plausible distractors on \textbf{model performance}, we group questions from the development set of {\dn} by the largest distractor plausibility (\ie, $\max_i \gamma_i$), in range of [$-1$, $1$], for each question and compare the performance of Co-Matching, BERT, and BERT-wwm-ext in different groups. As shown in Figure~\ref{fig:eval:distractorplaus}(a), the largest distractor plausibility may serve as an indicator of the difficulty level of questions presented to the investigated models. When the largest distractor plausibility is smaller than $-0.8$, all three models exhibit strong performance ($\geq90\%$). As the largest distractor plausibility increases, the performance of all models consistently drops. All models perform worse than average on questions having at least one high-plausible distractor (\eg, distractor plausibility $>0$). Compared with BERT, the gain of the best-performing model (\ie, BERT-wwm-ext) mainly comes from its superior performance on these ``difficult'' questions. %

Further, we find that distractor plausibility is strongly correlated with \textbf{the need for prior knowledge} when answering questions in {\dn} based on the annotated instances, as shown in Figure~\ref{fig:eval:distractorplaus}(b). For further analysis, we group annotated instances by different $\max_i\mathcal{S}(w_i,d)$ and $\mathcal{S}(c,d)$ (in Equation~(\ref{eq:dd})) and separately compare their need for linguistic knowledge and general world knowledge. As shown in Figure~\ref{fig:eval:distractorsim}, general world knowledge is crucial for question answering when the correct answer option is not mentioned explicitly in the document (\ie, $\mathcal{S}(c,d)$ is relatively small). In contrast, we tend to require linguistic knowledge when both the correct answer option and the most confusing distractor (\ie, the one with the largest distractor plausibility) are very similar to the given document.

\begin{figure}[ht]
   \begin{center}
   \includegraphics[width=0.38\textwidth]{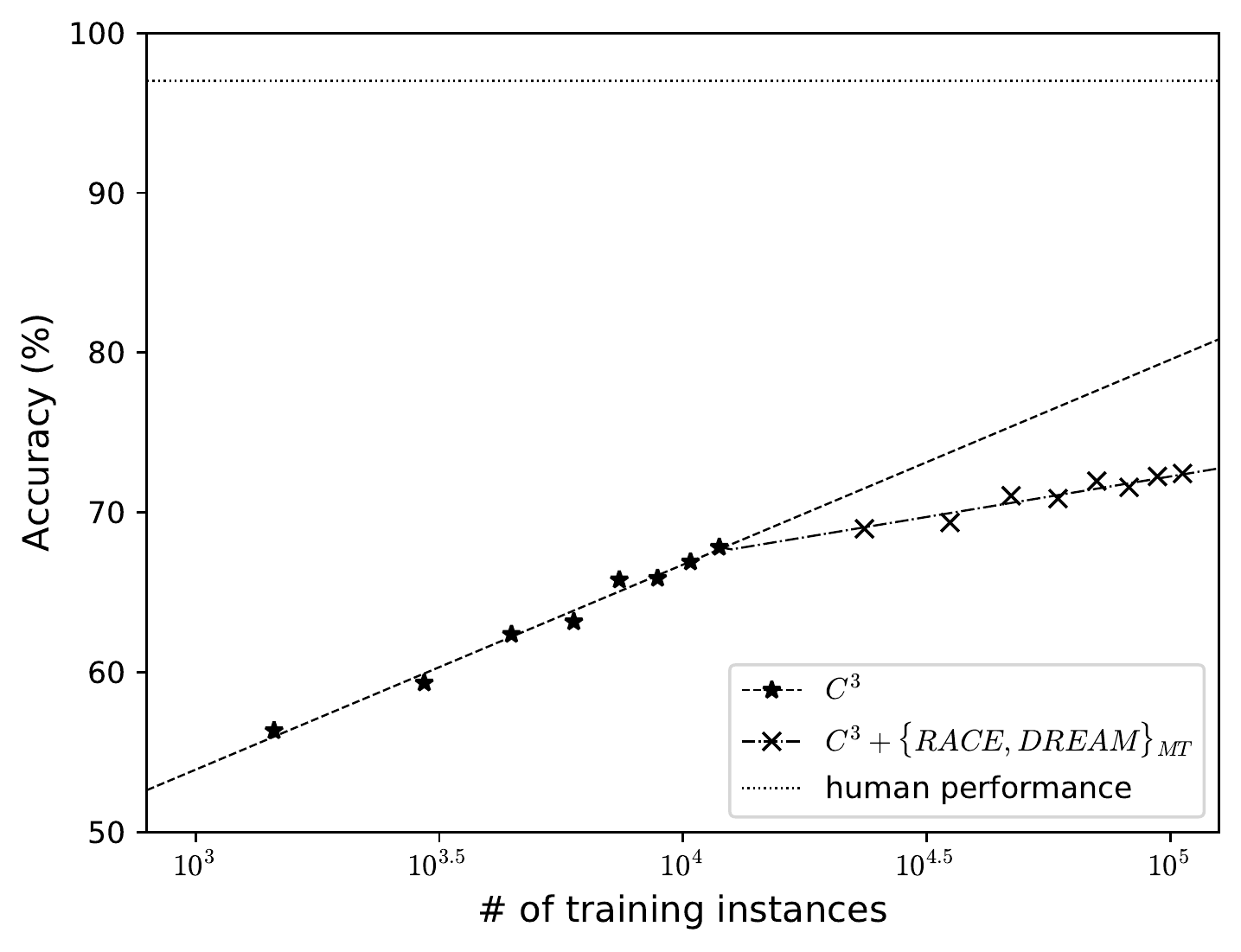}
   \end{center}
 \caption{Performance of BERT-wwm-ext trained on $1/8, 2/8, \ldots, 8/8$ of {\dn} training data, and {\dn} training data plus $1/8, 2/8, \ldots, 8/8$ of machine translated (MT) RACE and DREAM training data.}
 \label{fig:eval:nmt}
\end{figure}

\subsection{Discussions on Data Augmentation}

To extrapolate to what extent we can improve the performance of current models with more training data, we plot the development set performance of BERT-wwm-ext trained on different portions of the training data of {\dn}. As shown in Figure~\ref{fig:eval:nmt}, the accuracy grows roughly linearly with the logarithm of the size of training data, and we observe a substantial gap between human performance and the expected BERT-wwm-ext performance, even assuming that $10^5$ training instances are available, leaving much room for improvement. 

Furthermore, as the knowledge type distributions of {\dn} and its English counterparts RACE and DREAM are highly similar (Section~\ref{data:knowledge}), we translate RACE and DREAM from English to Chinese by Google Translate and plot the performance of BERT-wwm-ext trained on {\dn} plus different numbers of translated instances.
The learning curve is also roughly linear with the logarithm of the number of training instances from translated RACE and DREAM, but with a lower growth rate. Even augmenting the training data with all $94$k translated instances only leads to a $4.6\%$ improvement (from $67.8\%$ to $72.4\%$) in accuracy on the development set of {\dn}. 
From another perspective, BERT-wwm-ext trained on all translated instances \textbf{without} using any data in {\dn} only achieves an accuracy of $67.1\%$ on the development set of {\dn}, slightly worse than $67.8\%$ achieved when only the training data in {\dn} is used, whose size is roughly $1/8$ of that of the translated instances. These observations suggest a need to better leverage large-scale English resources from similar MRC tasks.

Besides augmenting the training data with translated instances, we also attempt to fine-tune a pre-trained \textbf{multilingual} BERT-Base released by \newcite{bert2018} on the training data of {\dn} and all \emph{original} training instances in English from RACE and DREAM. However, the accuracy on the development set of {\dn} is $63.4\%$, which is even lower than the performance ($65.7\%$ in Table~\ref{tab:eval:baseline}) of fine-tuning Chinese BERT-Base only on {\dn}.

\section{Conclusion}

We present the first free-form multiple-choice Chinese machine reading comprehension dataset ({\dn}), collected from real-world language exams, requiring linguistic, domain-specific, or general world knowledge to answer questions based on the given written or orally oriented texts. We study the prior knowledge needed in this challenging machine reading comprehension dataset and carefully investigate the impacts of distractor plausibility and data augmentation (based on similar resources for English) on the performance of state-of-the-art neural models. Experimental results demonstrate the there is still a significant performance gap between the best-performing model ($68.5\%$) and human readers ($96.0\%$) and a need for better ways for exploiting rich resources in other languages.

\section*{Acknowledgments}
We would like to thank the editors and anonymous reviewers for their helpful and insightful feedback.

\bibliography{tacl2018}
\bibliographystyle{acl_natbib}

\end{CJK*}

\end{document}